\documentclass{article}
\usepackage{spconf,amsmath,graphicx}
\usepackage{algorithmic}
\usepackage{algorithm}
\usepackage{amsfonts}
\usepackage{dsfont}

\title{Improving Visual Quality and Transferability of Adversarial Attacks on Face Recognition Simultaneously with Adversarial Restoration}

\name{Fengfan Zhou, Hefei Ling$^{\dagger}$\thanks{$^\dagger$ Corresponding Author (lhefei@hust.edu.cn). This work was supported in part by the Natural Science Foundation of China under Grant 61972169,62372203 and 62302186, in part by the National key research and development program of China(2022YFB2601802), in part by the Major Scientific and Technological Project of Hubei Province (2022BAA046, 2022BAA042), in part by the Knowledge Innovation Program of Wuhan-Basic Research, in part by China Postdoctoral Science Foundation 2022M711251.	
}, Yuxuan Shi, Jiazhong Chen, Ping Li}
\address{School of Computer Science and Technology, Huazhong University of Science and Technology, China}

\begin{document}
%
\maketitle

\begin{abstract}
Adversarial face examples possess two critical properties: Visual Quality and Transferability. However, existing approaches rarely address these properties simultaneously, leading to subpar results. To address this issue, we propose a novel adversarial attack technique known as Adversarial Restoration (AdvRestore), which enhances both visual quality and transferability of adversarial face examples by leveraging a face restoration prior. In our approach, we initially train a Restoration Latent Diffusion Model (RLDM) designed for face restoration. Subsequently, we employ the inference process of RLDM to generate adversarial face examples. The adversarial perturbations are applied to the intermediate features of RLDM. Additionally, by treating RLDM face restoration as a sibling task, the transferability of the generated adversarial face examples is further improved. Our experimental results validate the effectiveness of the proposed attack method.
\end{abstract}
\section{Introduction}
\label{sec:intro}
Due to the continuous advancements in deep learning, the performance of Face Recognition (FR) models has shown remarkable progress over time\cite{adaface}\cite{dcface}\cite{curricular_face}\cite{circle_loss}. Notably, for widely-used datasets like LFW\cite{lfw}, the performance of FR models has reached a saturation point\cite{arcface}\cite{cosface}\cite{dfanet}. Unfortunately, these models remain highly susceptible to adversarial examples\cite{dfanet}\cite{fim}, posing significant risks to the integrity of existing FR systems.
Consequently, there is a pressing need to enhance the performance of current adversarial face examples in order to expose potential vulnerabilities in FR models.

\begin{figure*}[htbp]
	\centering
	\includegraphics[width=120mm]{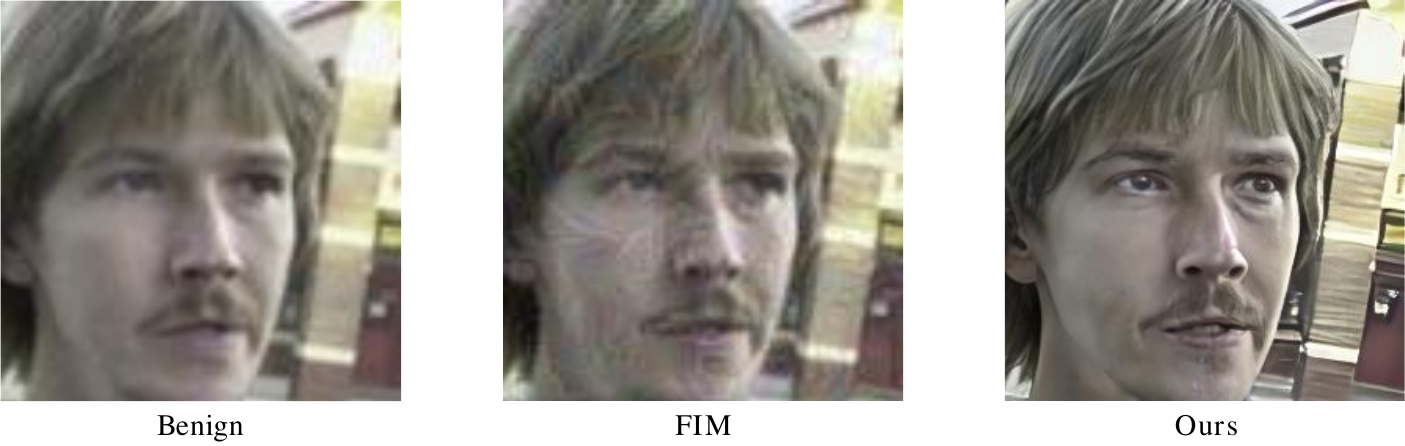}
	\caption{Illustration of the benign image, adversarial face example crafted by FIM\cite{fim}, and adversarial face example crafted by our proposed attack method.}
	\label{fig:adv_image_compare_26}
\end{figure*}

Adversarial face examples exhibit two crucial properties: Visual Quality and Transferability.
Unfortunately, existing adversarial attacks on FR fail to enhance both the visual quality and transferability of the generated adversarial face examples simultaneously, thereby limiting their potential applications.
To exacerbate the situation, the adversarial face examples generated by existing adversarial attacks on FR exhibit lower quality compared to the clean images before the addition of perturbations (see Fig.\ref{fig:adv_image_compare_26} for an example).
The aforementioned characteristic of adversarial face examples produced by existing adversarial attacks on FR presents an opportunity for human observers or adversarial detectors to potentially identify them. To address this, we draw inspiration from \cite{sibling_attack} and select the face restoration task as a sibling task. This strategic choice allows us to enhance both the visual quality and transferability of the generated adversarial face examples. By incorporating face restoration as a sibling task, we aim to increase both visual quality and transferability of the adversarial examples.

Specifically, we propose a diffusion-based face restoration model inspired by SR3\cite{sr3} and IDM\cite{idm}, termed as RLDM.
After RLDM is trained, we use the RLDM inference process to craft the adversarial face examples.
During the inference process of RLDM, we add adversarial perturbations on the output of the UNet of RLDM.

The experimental results demonstrate the effectiveness of our proposed method in simultaneously enhancing the visual quality and transferability of the crafted adversarial face examples.

\section{Methodology}
\label{sec:format}
Let $\mathcal{F}^{vct}(x)$ denote an FR model utilized by the victim for extracting the embedding vector from a face image $x$.
Let $x^{s}$ represent the image of the attacker, intended to target the FR system, and $x^t$ depict the victim image which the attacker aims to deceive.
The objective of the adversarial attacks in this paper is to fool $\mathcal{F}^{vct}$ into recognizing the adversarial face example $x^{adv}$ as $x^t$, while ensuring that $x^{adv}$ visually resembles $x^{s}$. Specifically, the objective can be expressed as follows:
\begin{equation}
	\begin{gathered}
		x^{adv}=\mathop{\arg\min}\limits_{x^{adv}}\left(\mathcal{D}\left(\mathcal{F}^{vct}\left(x^{adv}\right), \mathcal{F}^{vct}\left(x^{t}\right)\right)\right) \\
		\text{s.t.} \Vert x^{adv} - x^{s}\Vert_p \leq \varrho
		\label{eq:opt_obj}
	\end{gathered}
\end{equation}
where $\mathcal{D}$ is a pre-selected distance metric, and $\varrho$ is the maximum allowable perturbation magnitude.

In practical application scenarios, the attacker cannot direct access to the victim models, making it challenging to directly realize Eq.\ref{eq:opt_obj}. To this end, a commonly employed approach is to utilize a surrogate model, denoted as $\mathcal{F}$, to generate the adversarial examples. These generated adversarial examples are then transferred to the victim models for conducting attacks\cite{bpfa}. Therefore, the transferability of adversarial face examples becomes a critical factor in achieving successful attacks.

The loss function for crafting the adversarial face examples using the surrogate model $\mathcal{F}$ can be expressed as:
\begin{equation}
	\mathcal{L}=\Vert\phi\left(\mathcal{F}\left(x^{adv}\right)\right)-\phi\left(\mathcal{F}\left(x^{t}\right)\right)\Vert^2_2\label{eq:bpfa_loss}
\end{equation}
where $\phi(x)$ denotes the operation that normalizes $x$.

In addition to transferability, the visual quality of adversarial face examples plays a critical role in determining the effectiveness of adversarial attacks in real-world applications.
However, existing attack methods often result in a degradation of visual quality in crafted adversarial face examples (as depicted in Fig. \ref{fig:adv_image_compare_26}).
To improve the visual quality and transferability of the adversarial face examples simultaneously, we propose a novel attack approach called Adversarial Restoration (AdvRestore), which leverages a face restoration prior to enhance the visual quality and transferability of the crafted adversarial examples.
\begin{figure*}[htbp]
	\centering
	\includegraphics[width=140mm]{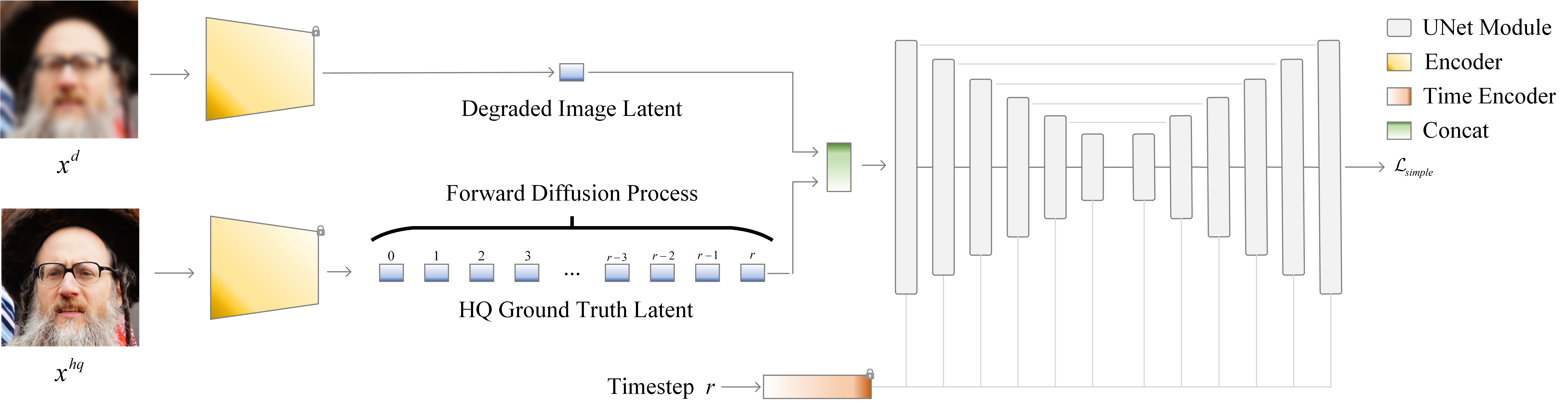}
	\caption{The framework of the proposed Restoration Latent Diffusion Model (RLDM).}
	\label{fig:rldm_framework}
\end{figure*}

Based on the inspiration from SR3\cite{sr3} and IDM\cite{idm}, we propose a novel end-to-end diffusion-based face restoration model called RLDM.
The framework of RLDM is presented in Fig. \ref{fig:rldm_framework}. The key distinction between our proposed RLDM and the approaches described in \cite{sr3}\cite{idm} lies in the utilization of image latent\cite{ldm} for restoring the degraded image.
This choice is motivated by the lower computational overhead associated with image latent usage and the enhancement in image quality, as demonstrated in \cite{ldm}.

RLDM incorporates a timestep schedule $T_r=\{1,2,...,N_r\}$. Let $B_r$ represents a predefined variance schedule $\beta_1, \beta_2,..., \beta_{N_r}$ corresponding to $T_r$. During the forward diffusion process, Gaussian noise is gradually added to the latent of $x^{hq}$ according to $B_r$. Conversely, in the reverse diffusion process, the latent corresponding to $x^{hq}$ is gradually denoised using the pretrained UNet model $\mathcal{U}$.
To accelerate the generation process of adversarial face examples, we employ DDIM\cite{ddim} to reduce the timesteps of the reverse diffusion process. Specifically, DDIM samples a subset of timesteps from $T_r$ and creates a new timestep schedule $S$. The denoising step in the reverse diffusion process of DDIM is performed solely using the sampled timesteps from $S$.
Let $\alpha_r=1-\beta_r$ and $\bar{\alpha}_r=\prod_{\kappa=1}^r \alpha_\kappa$.
To train RLDM, we start by acquiring the latent of $x^{d}$ and $x^{hq}$ using the Encoder from the LDM model\cite{ldm}:
\begin{equation}
	z^{d} = Enc\left(x^{d}\right)\label{get_latent_degraded_img}
\end{equation}
\begin{equation}
	z^{hq} = Enc\left(x^{hq}\right)
\end{equation}

Afterward, the forward process of LDM is executed $r$ times to introduce Gaussian noise onto $z^{hq}$. This procedure can be represented succinctly by the following formula:
\begin{equation}
	z^{hq}_{r}=\sqrt{\bar{\alpha}_r} z^{hq}_0+\sqrt{1-\bar{\alpha}_r} \xi
\end{equation}
where $z^{hq}_0=z^{hq}$, and $\xi \sim \mathcal{N}\left(0, \mathbf{I}\right)$

Drawing inspiration from SR3\cite{sr3} and IDM\cite{idm}, we fuse $z^{hq}_{r}$ and $z^{d}$ by concatenating them, forming a combined latent that we then feed into the UNet model. Additionally, along with the timestep embedding, we can obtain the output of the UNet within the RLDM model:
\begin{equation}
	\epsilon_\theta=\mathcal{U}\left(z^{d},z^{hq}_{r}, \mathcal{T}\left(r\right) \right)
\end{equation}
where $\mathcal{T}$ is the Time Encoder of RLDM.

We optimize UNet of the RLDM model using the following loss function:
\begin{equation}
L_{simple}=\mathbb{E}_{x, \epsilon \sim \mathcal{N}(0,1), r}\left[\left\|\epsilon-\epsilon_\theta\right\|_2^2\right]
\end{equation}
where $r$ is uniformly sampled from $T_r$.

During the training process of RLDM, we keep the parameters of the Encoder and Decoder fixed, while solely optimizing the parameters of the UNet.
Once RLDM is trained, we can generate adversarial face examples using the inference process of RLDM. This process is based on DDIM\cite{ddim} and involves several similar steps. In the subsequent discussion, we will focus on one particular step, noting that the complete inference process can be accomplished by repeating these similar steps multiple times.

To begin, we obtain the degraded image latent $z^{d}$ by applying the Encoder, as expressed in Eq. \ref{get_latent_degraded_img}. Subsequently, we input $z^{d}$ into the UNet $\mathcal{U}$ of RLDM, resulting in the following expression:
\begin{equation}
	\epsilon_\theta=\mathcal{U}\left(z^{d},z^n, \mathcal{T}\left(r\right) \right)
\end{equation}
where $z^n \sim \mathcal{N}\left(0,1\right)$.

Using $\epsilon_\theta$, we can then obtain the latent in the previous timestep, denoted as $z^{pre}$, by employing the function $\mathcal{H}$:
\begin{equation}
	z^{pre}=\mathcal{H}\left(\epsilon_\theta, z\right)
\end{equation}
\begin{equation}
	\mathcal{H}\left(\epsilon_\theta, z\right)=\sqrt{\bar{\alpha}_{r-1}} \tilde{z}+z^{\prime}+\sigma_r \epsilon^{\prime}
\end{equation}
where $\tilde{z}=\left(z-\sqrt{1-\bar{\alpha}_r} \epsilon_\theta\right) / \sqrt{\bar{\alpha}_r}$, $z'=\sqrt{1-\bar{\alpha}_{r-1}-\sigma_r^2} \epsilon_\theta$, $\sigma_r=\sqrt{\frac{\left(1-\bar{\alpha}_{r-1}\right)}{1-\bar{\alpha}_r}\left(1-\frac{\bar{\alpha}_r}{\bar{\alpha}_{r-1}}\right)}$, and $\epsilon' \sim \mathcal{N}\left(0, \mathbf{I}\right)$. 

Prior research has demonstrated that the reverse diffusion process can purify the added adversarial perturbations\cite{diffpure}. In order to circumvent this issue, we incorporate the adversarial perturbation onto the output feature of $\mathcal{U}$ when the timestep of the reverse diffusion process reaches one. This approach prevents the crafted adversarial perturbation from being eliminated through excessive reverse diffusion steps. Algorithm of our proposed AdvRestore when crafting adversarial face examples is demonstrated in Algorithm 1.

\begin{algorithm}
	\renewcommand{\algorithmicrequire}{\textbf{Input:}}
	\renewcommand{\algorithmicensure}{\textbf{Output:}}
	\caption{Adversarial Restoration Attack Method} 
	\label{alg:bpfa} 
	\begin{algorithmic}[1]
		\REQUIRE Negative face image pair $\{x^s, x^t\}$, the step size of the adversarial perturbations $\beta$, the maximum number of iterations $N_{max}$, maximum allowable perturbation magnitude $\varrho$, the DDIM timestep sequence $\mathcal{S}$, the function to get the latent in the previous timestep $\mathcal{H}$
		\ENSURE An adversarial face example $x^{adv}$
		\STATE $z^{s}=Enc\left(x^{s}\right), z \sim \mathcal{N}\left(0, \mathbf{I}\right)$, $m=\vert \mathcal{S} \vert$
		\FOR{$s=\mathcal{S}_m ,..., \mathcal{S}_1$}
		\STATE $\epsilon_\theta=\mathcal{U}\left(z^{s},z, \mathcal{T}\left(s\right) \right)$
		\STATE $z=\mathcal{H}\left(\epsilon_\theta, z\right)$
		\ENDFOR
		\STATE $\bar{x}=Dec\left(z\right)$
		\FOR{$t=1 ,..., N_{max}$}
		\STATE $\mathcal{L}=\Vert\phi\left(\mathcal{F}\left({\rm Clip}_{\bar{x}}^\varrho\left( Dec\left(z\right)\right)\right)\right)-\phi\left(\mathcal{F}\left(x^{t}\right)\right)\Vert^2_2$
		\STATE $\epsilon_\theta = \epsilon_\theta-\beta {\rm sign}\left(\nabla_{\epsilon_\theta}\mathcal{L}\right)$
		\STATE $z=\mathcal{H}\left(\epsilon_\theta, z\right)$
		\ENDFOR
		\STATE $x^{adv}={\rm Clip}_{\bar{x}}^\varrho \left(Dec\left(z\right)\right)$
	\end{algorithmic} 
\end{algorithm}

\section{Experiments}
\label{sec:pagestyle}
\subsection{Experimental setting}
\textbf{Datasets.}
For our experiments, we use LFW\cite{lfw} as the benchmark. LFW is a face dataset commonly employed in evaluating the performance of FR models in unconstrained scenarios, as well as assessing the effectiveness of adversarial attacks on FR. It is worth mentioning that the LFW dataset adopted in this paper is identical to the dataset utilized in \cite{bpfa}.

\textbf{Models.} In our experiments, we employed well-established FR models including FaceNet, MobileFace, IRSE50, and IR152. These models are the same as the models used in \cite{bpfa}\cite{adv_makeup}\cite{amt_gan}.
The encoder and decoder of RLDM model we used are the official LDM\cite{ldm} pretrained models for generating face images.

\textbf{Evaluation Metrics.} We use SSIM, PSNR, LPIPS, and Visual Quality Score (VQS) to evaluate the visual quality of various face images.
VQS is an evaluation metric that assesses the visual quality of a picture based on actual human feedback, providing a measurement that aligns more closely with real-world application scenarios.
Following the previous works, we utilize \textit{attack success rate} (ASR) to assess the efficacy of different attack methods. ASR quantifies the ratio of adversarial face examples that are successfully attacked.

\subsection{Experimental result}
\textbf{Evaluations on Visual Quality.} 
To assess the effectiveness of our proposed attack method in improving visual quality, we compute the SSIM, PSNR, and LPIPS metrics for both the benign image and the adversarial face examples according to previous works. Additionally, we obtain real human feedback on the visual quality of the images by engaging three volunteers to score them on a scale of 0 to 100, where higher scores indicate higher visual quality. The average results of this evaluation are presented in Table \ref{tab:visual_quality}. Fig. \ref{fig:adv_image_compare_26} illustrates a comparison between the benign image, an adversarial face example crafted using the FIM method\cite{fim}, and an adversarial face example generated using our proposed attack method.
\begin{table}[]
	\centering
	\small
	\begin{tabular}{c|c|c|c|c}
		\hline
		Image                 & VQS           & SSIM           & PSNR          & LPIPS          \\ \hline
		\textit{Benign}                & 60.7          & 0.471          & 24.2          & 0.432          \\ \hline
		FIM                   & 38.3          & 0.351          & 23.6          & 0.442          \\
		FIM+AdvRestore    & \textbf{76.7} & \textbf{0.755} & \textbf{30.1} & \textbf{0.064} \\ \hline
		DFANet                & 23.3          & 0.348          & 23.5          & 0.447          \\
		DFANet+AdvRestore & \textbf{80.0} & \textbf{0.737} & \textbf{29.7} & \textbf{0.076} \\ \hline
	\end{tabular}
	\caption{The visual quality of adversarial face examples crafted by our proposed method and the baseline. The first column represents the various image types in which \textit{Benign} means the attacker image without adding adversarial perturbations.}
	\label{tab:visual_quality}
\end{table}

Table \ref{tab:visual_quality} clearly demonstrates a significant improvement in the visual quality of the crafted adversarial face examples after the addition of AdvRestore to the alternate approaches.
The reason for this improvement is that our proposed AdvRestore enhances the visual quality of face images using our proposed RLDM. Importantly, the visual quality of the adversarial examples generated by our attack method surpasses that of the attacker images without any added adversarial perturbations, confirming the efficacy of adversarial perturbations in enhancing the visual quality of face images.

\begin{figure}[htbp]
	\centering
	\includegraphics[width=52mm]{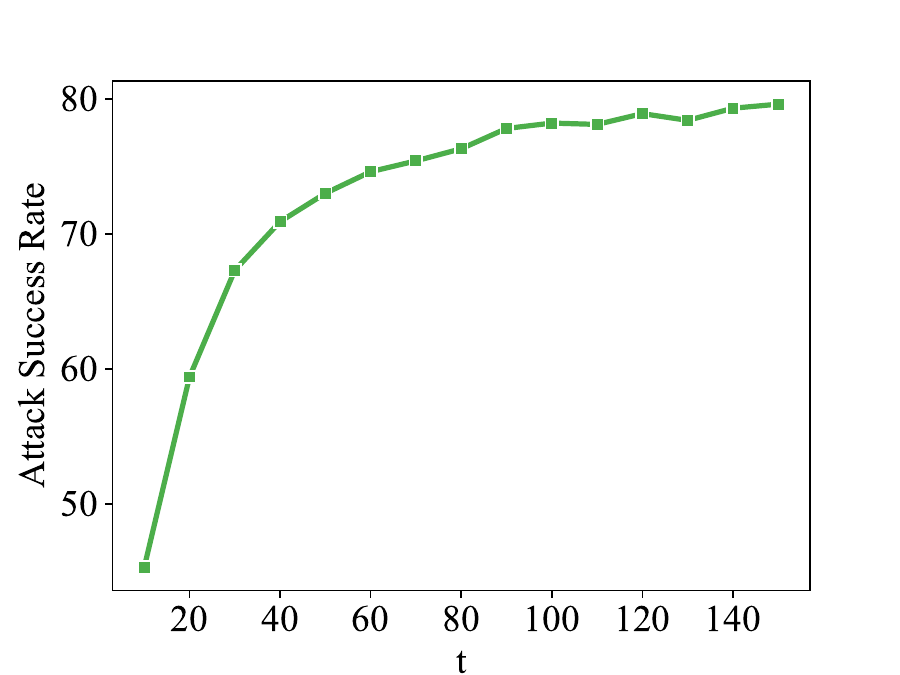}
	\caption{The adversarial success rate in different iterations.}
	\label{fig:adv_restore_asr_in_diff_round}
\end{figure}

\textbf{Evaluations on Attack Success Rate.}
We assess the effectiveness of our proposed attack method in enhancing the transferability of the crafted adversarial face examples by utilizing MobileFace as the surrogate model and generating adversarial face examples using various alternate attacks.
The results obtained on both normally trained models and adversarially robust models are presented in Table \ref{tab:asr_normal_trained_models} and Table \ref{tab:asr_adversarial_robust_models}, respectively. It is worth noting that the white-box ASR of all attacks is 100.0\%, hence we omit reporting the white-box ASR of the attacks in Table \ref{tab:asr_normal_trained_models} and Table \ref{tab:asr_adversarial_robust_models}.
To enhance the clarity of our experiment, we illustrate the ASR in different iterations in Fig.\ref{fig:adv_restore_asr_in_diff_round}.
\begin{table}[]
	\small
	\centering
	\begin{tabular}{c|c|c|c}
		\hline
		Attacks               & IR152        & IRSE50        & FaceNet       \\ \hline
		FIM                   & 4.4          & 65.4          & 6.7           \\
		FIM+AdvRestore    & \textbf{6.8} & \textbf{77.8} & \textbf{7.9}  \\ \hline
		DFANet                & 5.8          & 79.0          & 8.2           \\
		DFANet+AdvRestore & \textbf{8.0}   & \textbf{80.9} & \textbf{10.2} \\ \hline
	\end{tabular}
	\caption{The attack success rates on LFW with normal trained models as the victim models. The first column demonstrates the attacks. The second to forth columns in the first row demonstrate the victim models.
}
	\label{tab:asr_normal_trained_models}
\end{table}

\begin{table}[]
	\centering
	\small
	\begin{tabular}{c|c|c|c}
		\hline
		& IR152$_{adv}$     & IRSE50$_{adv}$    & FaceNet$_{adv}$   \\ \hline
		FIM                   & 1.6          & 3.8          & 2.1          \\
		FIM+AdvRestore    & \textbf{3.1} & \textbf{7.5} & \textbf{3.5} \\ \hline
		DFANet                & 2.4          & 7.3          & 2.8          \\
		DFANet+AdvRestore & \textbf{3.7} & \textbf{8.4} & \textbf{4.7} \\ \hline
	\end{tabular}
\caption{The attack success rates on LFW with adversarial robust models as the victim models. The first column demonstrates the attacks. The second to forth columns in the first row demonstrate the victim models.}
\label{tab:asr_adversarial_robust_models}
\end{table}

Table \ref{tab:asr_normal_trained_models} and Table \ref{tab:asr_adversarial_robust_models} clearly illustrate a improvement in the black-box attack success rate after incorporating our proposed method. This improvement can be attributed to our attack method effectively leveraging the prior knowledge of the RLDM restoration sibling task, thereby enhancing the transferability of the crafted adversarial face examples.

\section{RELATION TO PRIOR WORK}
Our proposed method aims to enhance both the visual quality and transferability of crafted adversarial face examples. The inspiration for this work stems from the Sibling-Attack\cite{sibling_attack}, which has demonstrated that attribute recognition tasks can improve the transferability of adversarial face examples. However, the effectiveness of using face restoration tasks to enhance transferability, as well as the visual quality, has not been explored in \cite{sibling_attack}. Notably, the visual quality of adversarial examples crafted in \cite{sibling_attack} is lower than that of benign images without adding adversarial perturbations. In this paper, we address these issues by investigating the impact of employing a face restoration latent diffusion model to improve the transferability of adversarial face examples, while simultaneously enhancing their visual quality.
\section{Conclusion}
\label{sec:prior}
In this paper, we introduce a novel adversarial attack on FR, called AdvRestore, aimed at enhancing both the visual quality and transferability of adversarial face examples simultaneously. Through experimental evaluations, we demonstrate the effectiveness of our proposed attack method. Additionally, our results confirm that adversarial perturbations can indeed improve the visual quality of crafted face images rather than diminishing it.
\vfill\pagebreak
\bibliographystyle{IEEEbib}
\bibliography{strings,refs}

\end{document}